\documentclass[letterpaper]{article} 
\usepackage{aaai23}  
\usepackage{times}  
\usepackage{helvet}  
\usepackage{courier}  
\usepackage[hyphens]{url}  
\usepackage{graphicx} 
\usepackage{natbib}  
\usepackage{caption} 
\usepackage{algorithm}
\usepackage{algorithmic}
\usepackage{amsmath,amsfonts,amssymb}
\usepackage{siunitx}
\usepackage{multirow}
\usepackage{calc}
\newlength\myheight
\newlength\mydepth
\settototalheight\myheight{Xygp}
\newcommand*\inlinegraphics[1]{%
  \settototalheight\myheight{Xygp}%
  \settodepth\mydepth{Xygp}%
  \raisebox{-\mydepth}{\includegraphics[height=\myheight]{#1}}%
}

\usepackage{newfloat}
\usepackage{listings}
\DeclareCaptionStyle{ruled}{labelfont=normalfont,labelsep=colon,strut=off} 
\floatstyle{ruled}
\newfloat{listing}{tb}{lst}{}
\floatname{listing}{Listing}
\title{Safety Shielding under Delayed Observation}
\author{
    Filip Cano C\'ordoba\textsuperscript{\rm 1},
    Alexander Palmisano\textsuperscript{\rm 1},
    Martin Fr\"anzle\textsuperscript{\rm 2},\\
    Roderick Bloem\textsuperscript{\rm 1},
    Bettina K\"onighofer\textsuperscript{\rm 1}
}
\affiliations{
    \textsuperscript{\rm 1} Institute of Applied Information Processing and Communications, Graz University of Technology\\
    \textsuperscript{\rm 2} Dpt.\ of Computing Science, Carl von Ossietzky University of Oldenburg\\
    filip.cano@iaik.tugraz.at, alexander.palmisano@student.tugraz.at, martin.fraenzle@uni-oldenburg.de, roderick.bloem@iaik.tugraz.at, bettina.koenighofer@iaik.tugraz.at
}

\begin{document}

\maketitle

\begin{abstract}
Agents operating in physical environments need to be able to handle delays in the input and output signals since neither data transmission nor sensing or actuating the environment are instantaneous. Shields are correct-by-construction runtime enforcers that guarantee safe execution by correcting any action that may cause a violation of a formal safety specification. Besides providing safety guarantees, shields should interfere minimally with the agent. Therefore, shields should pick the safe corrective actions in such a way that future interferences are most likely minimized. Current shielding approaches do not consider possible delays in the input signals in their safety analyses. In this paper, we address this issue. We propose synthesis algorithms to compute \emph{delay-resilient shields} that guarantee safety under worst-case assumptions on the delays of the input signals. We also introduce novel heuristics for deciding between multiple corrective actions, designed to minimize future shield interferences caused by delays. As a further contribution, we present the first integration of shields in a realistic driving simulator. We implemented our delayed shields in the driving simulator \textsc{Carla}. We shield potentially unsafe autonomous driving agents in different safety-critical scenarios and show the effect of delays on the safety analysis.
\end{abstract}

\section{Introduction} 
\label{sec:intro}

Due to the complexity of nowadays autonomous, AI-based systems, approaches that guarantee safety during runtime are gaining more and more attention~\cite{DBLP:conf/birthday/KonighoferBEP22}.
A maximally-permissive enforcer, often called a \emph{shield}, overwrites any actions from the agent that may cause a safety violation in the future~\cite{AlshiekhBEKNT18}.
In order to enforce safety while being maximally permissive, 
the shield has to compute the latest point in time where safety can still be enforced.
For that reason, shields are often computed by constructing a \emph{safety game} from
an environmental model that captures all safety-relevant dynamics and a formal safety specification. 
The \emph{maximally-permissive winning strategy $\rho$} 
allows, 
within any state, all actions that will not cause a safety violation over the infinite time horizon. Given a state, we call an action \emph{safe} if the action is contained in $\rho$, and an action is called \emph{unsafe} otherwise.
Shields allow any actions that are safe according to $\rho$.

Incorporating delays in safety computations is necessary for almost any real-world control problem. Delays are caused by data collection, processing, or transmission
and are therefore omnipresent for any agent operating in a complex environment. 
Not addressing these delays can be the root of many safety-critical problems. 
\begin{figure}[t]
    \centering
    \includegraphics[width=0.45\textwidth]{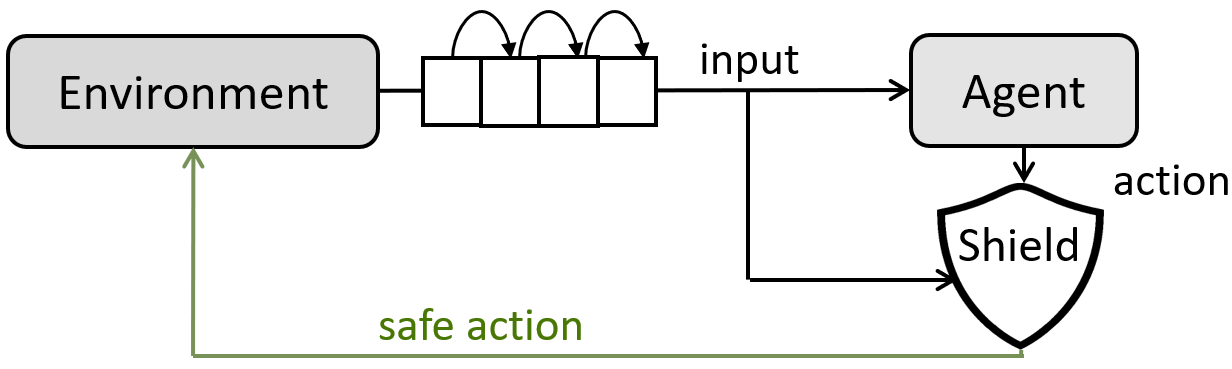}
    \caption{Delay-resilient shielding scheme.}
    \label{fig:shielding_setting}
\end{figure} 

\emph{Example. 
Let us assume that a car detects a pedestrian at position $(x,y)$, 
and it is aware of a time delay $\delta$ between sensing and acting. 
The vehicle has to plan its next actions in such a way that they are safe for any position of the pedestrian in the interval $(x\pm \varepsilon, y \pm \varepsilon)$, where $\varepsilon$ is defined via assumptions on the pedestrian's velocity and the delay $\delta$.}

In this paper, we propose synthesis algorithms for \emph{delay-resilient} shields, 
i.e. shields that guarantee safety under assumptions on the worst-case delay on the inputs. 
Figure~\ref{fig:shielding_setting} shows the shielding settings under delay.
 
To synthesize delay-resilient shields, we incorporate a worst-case delay in the safety game, which induces imperfect state information~\cite{Chen2020IndecisionAD}. 
The delay-resilient shields are then computed from the maximally-permissive winning strategy in the delayed safety game.
In order to obtain a fixed replacement action for any unsafe action,
we have to determinize the maximally-permissive strategy.
To do so, we can define a property over the state space
and set the action maximizing such property as the one fixed by the shield.
We study two such properties: controllability and robustness.
The \emph{controllability value} assigns to any state $s$ the \emph{maximal delay} on the input under which $s$ stays safe. 
The \emph{robustness value} of a state $s$ is the length of the minimal path from $s$ to any unsafe state.
We discuss how to maximize a state property under the uncertainty introduced by the delayed input.

In our experiments, we integrate shielding under delay in the driving simulator 
\textsc{Carla}~\cite{dosovitskiy_carla_2017}.
Our results show the effects of delays on the safety analysis and that our method is scalable enough to be applied in complex application domains.
As a second case study, we perform experiments on a gridworld
and compare the performance of delay-resilient shields with different worst-case delay.
The source code to reproduce the experiments, 
are available on the accompaning repository\footnote{\url{https://github.com/filipcano/safety-shields-delayed}}.

\paragraph{Related work.}
Shields for discrete systems were introduced in \cite{DBLP:conf/tacas/BloemKKW15}
and several extensions and applications have already been published~\cite{DBLP:conf/isola/TapplerPKMBL22,DBLP:conf/amcc/PrangerKTD0B21,0001KJSB20}, e.g., shielding for reinforcement learning agents~\cite{Carr2022, DBLP:conf/nfm/KonighoferRPTB21,DBLP:conf/atal/Elsayed-AlyBAET21}. 
Chen et al.~\cite{chen2018s,Chen2020IndecisionAD} investigated the synthesis problem for time-delay discrete systems by the reduction to solving 
two-player safety games. 
We base our shields 
on their proposed algorithm for solving delayed safety games. 
Note that the delayed games discussed in~\cite{DBLP:journals/iandc/WinterZ20}
follow a different concept.
In their setting, \emph{a delay is a lookahead} granted by the input player as an advantage to the delayed player: the delayed player P1 lags behind input
player P0 in that P1 has to produce the $i$-th action when
$i+j$ inputs are available. 
In contrast, 
we do not grant a lookahead into future inputs but consider reduced information due to input data being delivered to the agent with delay, which renders our agent equivalent to their input player. 
The notion of delay employed in this paper also is different from that in timed games~\cite{DBLP:conf/cav/BehrmannCDFLL07}.
In timed games, delay refers to the possibility of deliberately
delaying the next single action. 
However, both players have full and up-to-date information in timed games. 
In the continuous and hybrid domains, control barrier functions~\cite{DBLP:conf/eucc/AmesCENST19} are used to enforce safety.
Prajna et al. extended the notion of barrier certificates to time-delay systems~\cite{Prajna05}.
Bai et al.~\cite{BaiGJX0Z21} introduced a new model of hybrid
systems, called delay hybrid automata, to capture the continuous dynamics of dynamical
systems with delays. 
However, this work does not address the fact that 
state observation in embedded systems is \textit{de facto} in discrete time 
and that a continuous-time shielding mechanism therefore would require adequate interpolation between sampling points.

\section{Preliminaries - Shielding without Delays} 
\label{sec:shield_synthesis}
We briefly outline the classical approach for computing shields via safety games. 
We refer to~\cite{AlshiekhBEKNT18} for more details and formal definitions.
The classical approach to computing shields consists of the following steps:

\paragraph{Step 1. Construct the safety game.}
The possible interactions between the environment and the agent can naturally be modelled as a 2-player game. The game is played in alternating moves by the two
players: the environment player picks a next input $i \in \mathcal{I}$ 
(e.g., sensor data, movement of other agents), 
and the agent player picks a next action $a \in \mathcal{A}$.
The game is played on a game graph $\mathcal{G} = \langle \mathcal{S}, \mathcal{E} \rangle$.
The set $\mathcal{S}$ represents the states of the environment, including the state information of the agent that is operating within the environment. The transitions $\mathcal{E} : \mathcal{S} \times \mathcal{I} \times \mathcal{A} \rightarrow\mathcal{S} $ model how the states are updated, depending on the chosen input $i\in \mathcal{I}$ and the chosen action $a \in \mathcal{A}$.
The game graph is complemented by a \emph{winning condition} in form of a \emph{safety specification} which defines \emph{unsafe states} 
on  $\mathcal{G}$. The agent loses whenever the play reaches some unsafe state. 

\paragraph{Step 2. Compute the maximally-permissive winning strategy.}
The objective of the agent player is to always select actions avoiding
unsafe states, while the environment player tries to drive the game to an unsafe state by picking adequate inputs.
Solving the safety game refers to computing a winning strategy $\rho$ for the agent: any play that is played according to $\rho$ (i.e., the agent always picks actions that are contained in $\rho$)
is winning for the agent, meaning that no unsafe states are visited. 
For safety games with full information, memoryless winning strategies 
$\rho : \mathcal{S} \times \mathcal{I} \rightarrow{2^\mathcal{A}}$
exist~\cite{Thomas1995}.
A \emph{maximally-permissive} winning strategy subsumes the behaviour of every winning strategy, i.e., at any move, the maximally-permissive winning strategy allows any action that is contained in some winning strategy. 

\paragraph{Step 3. Implement a shield by fixing actions.}
For any move, we call an action \emph{safe} if the action is contained in the maximally-permissive winning strategy $\rho$, and call it \emph{unsafe} otherwise. To implement a shield, we have to define for every unsafe action a concrete safe replacement action. 

Given a state $s \in \mathcal{S}$, an input $i \in \mathcal{I}$, and an action $a \in \mathcal{A}$,
a shield is implemented in the following way:
\begin{itemize}
    \item If $a\in\rho(s,i)$, the shields outputs $a$.
    \item If $a\notin\rho(s,i)$, the shield outputs
    $a' \in \mathcal{A}$ with $a'\in\rho(s,i)$.
\end{itemize}
The shield is attached to the agent. At every time step, the shield reads the current input and suggested action from the agent, and either forwards the suggested action to the environment if it is safe ($a\in\rho(s,i)$), or replaces the action with a safe action $a'$. Different heuristics have been proposed to decide the choice of $a'$, all with the goal to minimize future shield interferences~\cite{DBLP:journals/fmsd/KonighoferABHKT17}.

\paragraph{Complexity.} Creating and solving the safety game (steps 1 and 2) 
has a cost of $\mathcal O(|\mathcal S|)$.
The cost of step 3 depends on the heuristic chosen to decide the corrective safe action.

\section{Shielding under Delayed Inputs} 

The setting for shielding under delayed inputs is depicted in Figure~\ref{fig:shielding_setting}.
The delayed information is forwarded sequentially
from the environment to the agent and to the shield.
This corresponds to having a FIFO-buffer
in the information channels.
Let us assume a worst-case delay of $\delta \in \mathbb{N}$ steps. 
The shield would therefore have to decide about the safety of an action after some finite execution
$s_0,i_1,a_1,s_1,i_2,a_2,\dots,s_n,i_n$ already having just seen its proper prefix $s_0,i_1,a_1,s_1,\dots,s_{n-\delta}$.
Thus, the shield is not aware of the current state $s_n$.
Instead, it only has access to a proper prefix
of the full state history. 
Nevertheless, the shield has to
decide on the safety of the current action $a_n$ of the agent without knowing the remainder of the state history.

\paragraph{Synthesis of delay-resilient shields.}
We propose a synthesis algorithm to compute delay-resilient shields. Our algorithm extends the classical game-based  synthesis approach by computing winning strategies under delay. 
Our synthesis algorithm performs the following steps:

\paragraph{Step 1-2. As for the delay-free case.}
Our algorithm to compute delay-resilient
shields starts by synthesizing a maximally-permissive winning strategy $\rho$ for the delay-free safety game $\mathcal{G}$, as discussed in the previous section.

\paragraph{Step 3. Compute winning strategy under delay.}
Playing a game under delay $\delta$ amounts to pre-deciding actions $\delta$ steps in advance.
Even though this makes the control problem harder,
the existence of a winning strategy under such delays is still decidable.
However, for games with delayed inputs, memoryless strategies are not powerful enough.
For a game under delay $\delta$, a winning strategy $\rho_{\delta}$ requires a memory of size $\delta$ to queue the $\delta$ latent actions, i.e.,
$\rho_{\delta} : \mathcal{S} \times \mathcal{I} \times \mathcal{A} ^\delta \rightarrow \mathcal{S}$.
Since straightforward reductions to delay-free games induce a blow-up of the game graph, 
which is strictly exponential in the magnitude of the delay~\cite{Tripakis04}, 
we use an 
incremental approach. 
In the following, we sketch the idea of the algorithm,
further details are in~\cite{Chen2020IndecisionAD}. 
The algorithm incrementally computes the maximally-permissive winning strategies for increasing delays and reduces game-graph size in between.
As controllability (i.e., the agent wins from this state) under
delay $k$ is a necessary condition for controllability under delay $k' > k$, each
state uncontrollable under delay $k$ can be removed before computing the winning strategy for larger delay.
The algorithm returns the maximally-permissive winning strategy 
$\rho_{\delta}$ that is winning the original game $\mathcal{G}$ under dely $\delta$.
Although the theoretical worst-case complexity is  $\mathcal O(|\mathcal S|^\delta)$, the incremental algorithm has been proven to be very efficient in practice~\cite{Chen2020IndecisionAD}.

\paragraph{Step 4. Implement a shield by fixing actions.}
A delay-resilient shield has to
correct actions that are unsafe under delay.
Given a state $s \in \mathcal{S}$, an input $i \in \mathcal{I}$, the $\delta$ latent actions
$A = [a_1,\dots,a_{\delta}] \in \mathcal{A}^{\delta}$, and the next action $a$,
a delay-resilient shield is implemented as follows:
\begin{itemize}
    \item If $a\in\rho_{\delta}(s,i,A)$, the shields outputs $a$.
    \item If $a\notin\rho_{\delta}(s,i,A)$, the shield outputs
    $a'\in\rho_\delta(s,i,A)$.
\end{itemize}

We propose two novel state properties used to 
decide on the concrete corrective action $a'$ selected in the delayed case.
\begin{enumerate}
    \item \emph{Controllability} $\phi_c$: 
    The value $\phi_c(s)$ of a state $s$ is \emph{the maximum delay for which $s$ is controllable}, using some threshold 
    $\delta_{\max}$ to limit the largest considered delay.
    \item \emph{Robustness} $\phi_r$:  The value $\phi_r(s)$ of a state $s$ is the length of the \emph{minimal path} from $s$ to any \emph{unsafe state}.
\end{enumerate}

Using $\phi_c$ as decision heuristic results 
in shields that minimize expected shield interferences caused by delays.
We compute the controllability values 
by computing the maximally-permissive winning strategies 
$\rho_1,\dots,\rho_{\delta_{\max}}$ for the delays 
$\delta\in\{1,\dots,\delta_{\max}\}$ and using them to decide on the controllability of states.
The cost of pre-computing this heuristic for all states is the cost of solving the game for delay up to $\Delta$, in the worst-case $\mathcal O(|\mathcal S|^{\delta_{\max}})$.

Using $\phi_r$ as decision heuristic results 
in shields that minimize future expected shield interferences caused by actions that violate safety, 
not necessarily related to safety violations due to delays.
Intuitively, this is the case since a high robustness value suggests that the agent is in a state that ``easily'' satisfies safety, 
while values near zero suggest that the system is close to violating it.
The cost of computing this heuristic for all states is $\mathcal O(|S|)$, which adds to the time for solving the delayed safety game.

While the decision heuristic of $\phi_c$ is specially designed for minimizing shield interferences caused by delays, computing  $\phi_c$ for large thresholds $\delta_{\max}$ is computationally expensive. In the experimental section, we will discuss that using shields maximizing $\phi_r$ resulted in almost the same interference rates while being less computationally expensive.

In the delayed setting, the shield is not aware of the current state.
Therefore, a delay-resilient shield has to pick a corrective action such that the \emph{expected}
controllability or robustness value is maximized.
Let $\delta$ be the worst-case delay, 
let $s$ be the last state the shield is aware of, 
and let $A= [a_1,\dots,a_{\delta}]$ be the buffer of latent actions. 
Then the \emph{forward set}
$\mathcal{S}_F \subseteq \mathcal{S}$ contains all states that can be reached from $s$ 
performing the actions $A$.
In other words, $s_f$ is contained in $\mathcal{S}_F$ if there exists a set of inputs $i_1,\dots,i_{\delta}\in \mathcal I$ such that
the execution defined by the transition relation of $\mathcal{G}$ is 
$s,i_1,a_1,\dots,i_{\delta},a_{\delta},s_f$.
We suggest picking the corrective action such that
the \emph{average} controllability or robustness value of the corresponding forward set is maximized. 

\section{Experiments - Shielded Driving in \textsc{Carla}} 
\label{sec:experiments}

We implemented our delayed shields in the driving simulator \textsc{Carla}~\cite{dosovitskiy_carla_2017}. 
In all scenarios, the default autonomous driver agent in \textsc{Carla} is used 
with adequate modifications to make it a more reckless driver.
To capture the continuous 
dynamics of \textsc{Carla}
using discrete models, 
we designed the safety game with 
overly conservative transitions, i.e., accelerations are overestimated and braking power is underestimated.
In both scenarios we use delay-resilient shields maximizing robustness.
All experiments were executed on a computer with
AMD Ryzen 9 5900 CPU, 32GB of RAM running
Ubuntu 20.04.

\subsection{Shielding against Collisions with Cars}

\begin{figure}[t]
  \centering
    \includegraphics[height=0.142\textwidth]{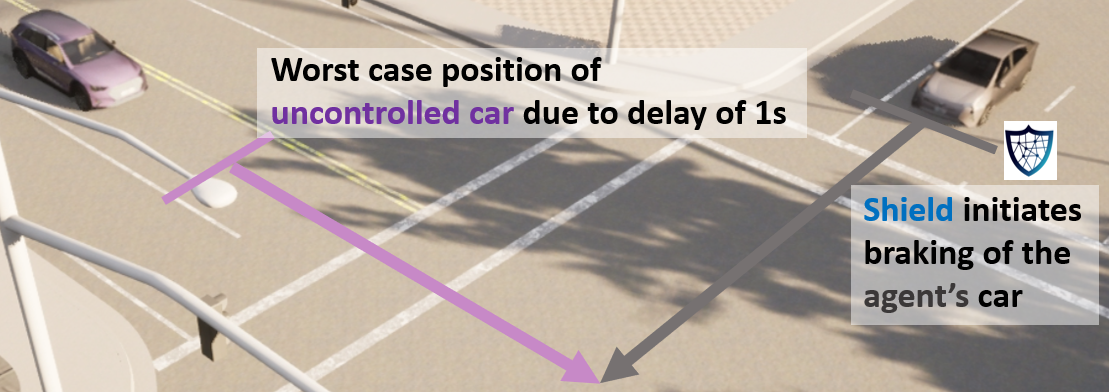}
    \caption{Scenario: cars at an intersection.}
    \label{fig:carscreenshot}
\end{figure}

We consider a scenario in which two cars (one of them controlled by the driver agent) approach an uncontrolled intersection. The shield has to guarantee collision avoidance for any braking and acceleration behaviour of the uncontrolled car, while the observation of the uncontrolled car is delayed. A screenshot of the \textsc{Carla} simulation is given in Figure~\ref{fig:carscreenshot}.

\paragraph{Shield computation.}
To compute delay-resilient shields, the 
scenario is encoded as a safety game $\mathcal{G} = \langle \mathcal{S}, \mathcal{E} \rangle$.
We model each car with two state variables:
\begin{itemize}
    \item $P_{\mathrm{agent}}$ and $P_{\mathrm{env}}$ represent respectively the distances of the agent's car and the environment's car to the crossing.
    The range is $P_\mathrm{agent} = P_\mathrm{env} = \{0, 2, 4, \dots, 100 \}~\unit{\metre}$.
    \item $V_{\mathrm{agent}}$ and $V_{\mathrm{env}}$ represent  the velocity of the agent’s car and the environment's car, resp.
    The range is $V_\mathrm{agent} = V_\mathrm{env} = \{0, 1, 2, \dots, 20 \}~\unit[per-mode=symbol]{\metre\per\second}$.
\end{itemize}
Each time step in the game corresponds to $\Delta t= 0.5~\unit{\second}$ in the simulation.
Each car can perform three actions: 
$\mathtt a$ (accelerate), $\mathtt b$ (brake) or $\mathtt c$ (coast, touch no pedal). 
Therefore, the set of inputs is
$\mathcal{I}=\{\mathtt a_\mathrm{env}, 
\mathtt b_\mathrm{env}, \mathtt c_\mathrm{env}\}$ and the set of actions is
$\mathcal{A}=\{\mathtt a_\mathrm{agent}, 
\mathtt b_\mathrm{agent}, \mathtt c_\mathrm{agent}\}$.
In our model, braking and throttling have the effect of applying a constant acceleration of $a=\pm 2~\unit[per-mode=symbol]{\metre\per\second^2}$.
Therefore, the position $p_t$ at time step $t$ is updated
\footnote{The velocity is applied as negative because the car
gets closer to the intersection at every step.}
as
$p_{t+\Delta t} = p_t - v_t\Delta t - \tfrac{1}{2}a\Delta t^2$, and the velocity as $v_{t+\Delta t} = v_t + a\Delta t$.
Unsafe states represent collisions, therefore
$\mathcal{S_\mathrm{unsafe}}=\{(p_\mathrm{agent}, v_\mathrm{agent},
p_\mathrm{env}, v_\mathrm{env}) : p_\mathrm{agent} = p_\mathrm{env}\}$.
From the safety game, we compute delay-resilient shields that maximize the expected robustness.

\begin{figure}[t]
    \centering
    \includegraphics[width=0.40\textwidth]
    {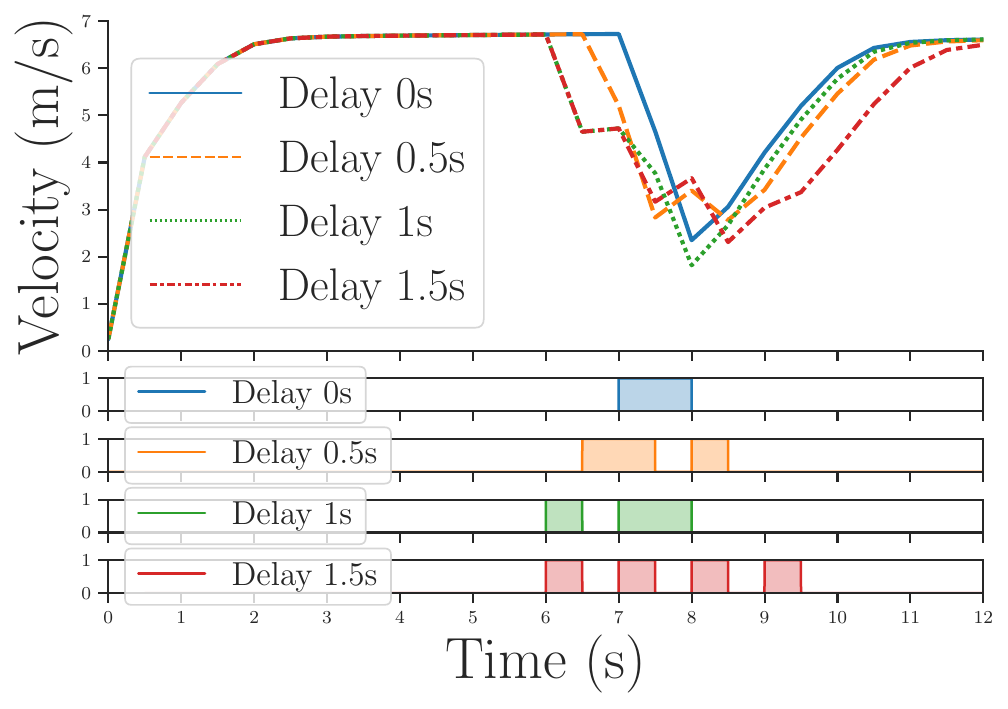}
    \caption{Velocity and shield activation over time.}
    \label{fig:car_results}
\end{figure}

\paragraph{Results.}
In Figure~\ref{fig:car_results}, we plot the speed of the agent's
car against time and shield interferences (coloured bars) for different delays,
expressed in steps of $\Delta t = 0.5~\unit{\second}$. 
As expected, the shield interferes over a longer time for increasing delays.
For delay 0, the agent's car brakes continuously until it escapes danger. 
For larger delays, the shields force the car to brake earlier, accounting for the worst-case behaviour of the other car. 
The shield always prepares for worst-case behaviour of the environment, 
which often does not materialize subsequently.
This explains why the shields change between activity and inactivity several times in the same execution, especially for larger delays.
We tested the shields for several safety-critical scenarios, varying positions and velocities, and were able to avoid collisions in all cases. 
In Table~\ref{tab:computingtimes}, we give the synthesis times to compute the shields.
Each delay step in Table~\ref{tab:computingtimes} corresponds to 
$\Delta t = 0.5~\unit{\second}$.

\begin{table}[b]
\begin{tabular}{llllll}
\multicolumn{2}{l}{Delay (in steps)}                                & 0    & 1   & 2     & 3     \\ \hline
\multirow{2}{*}{Syn. times (in s)}                & Car example   & 
1.5    & 13   & 48   & 167   \\ \cline{2-6} 
                                      & Pedestrian  ex. 
& 0.8  & 9 & 34     & 119 
\end{tabular}
\caption{Shield synthesis times (in seconds).}
\label{tab:computingtimes}
\end{table}

\subsection{Shielding against Collisions with Pedestrians}
\begin{figure}[t]
    \centering
    \includegraphics[height=0.142\textwidth]
    {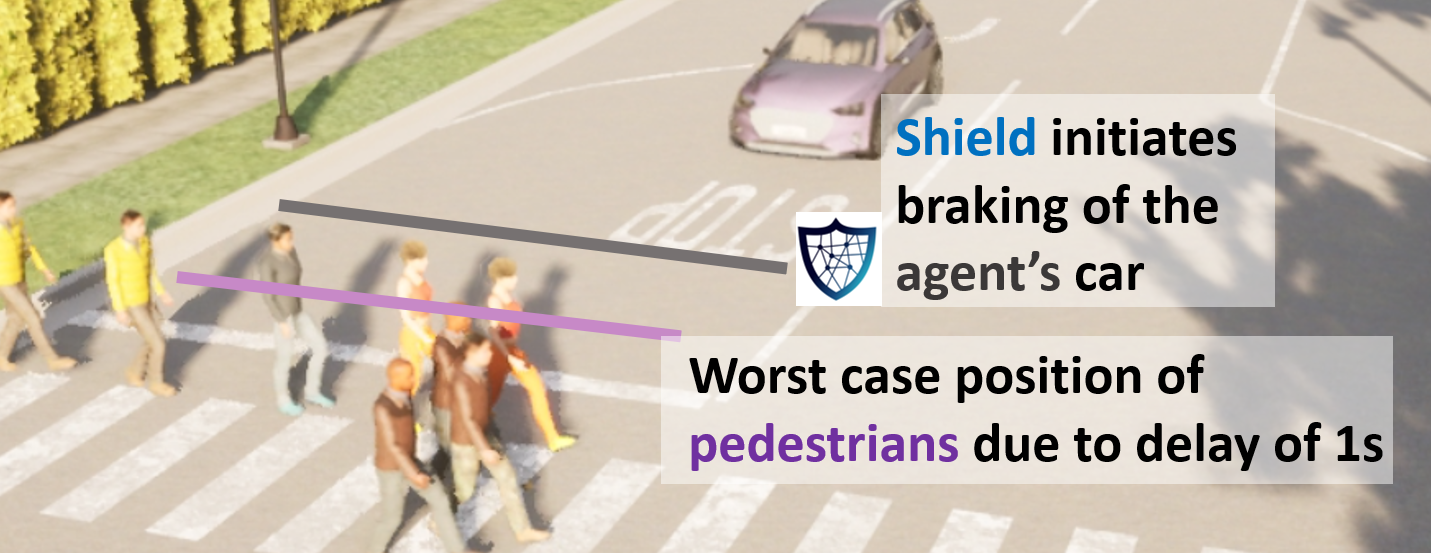}
    \caption{Scenario: pedestrians at crosswalk.}
    \label{fig:carlascreenshot}
\end{figure}

In the second experiment, we compute shields for collision avoidance with pedestrians. Similar to before, the shields guarantee safety under delay even under the worst possible behaviour of the pedestrians. 
A screenshot of the \textsc{Carla} simulation is given in Figure~\ref{fig:carlascreenshot}.

\paragraph{Shield computation.}
The car, which is controlled by the driver agent, is modelled in the same manner as before. Pedestrians are controlled by the environment and only have as state variables their position. 
In our model, we assume that a pedestrian can move $1~\unit{\metre}$ in any direction within one timestep of $\Delta t=0.5~\unit{\second}$.
We consider a state to be unsafe whenever the ego car moves fast while being close to a pedestrian and the pedestrian is closer to the crosswalk than the car.
Formally
$\mathcal{S_\mathrm{unsafe}}=\{(p_\mathrm{agent}, v_\mathrm{agent},
p_\mathrm{ped}) : 
(v_\mathrm{agent} > 2~\unit[per-mode=symbol]{\metre\per\second}
\land |p_\mathrm{agent}-p_\mathrm{ped}|<5~\unit{\metre}
\land p_\mathrm{ped} < p_\mathrm{agent})\}$.

\begin{figure}[t]
    \centering
    \includegraphics[width=0.48\textwidth]{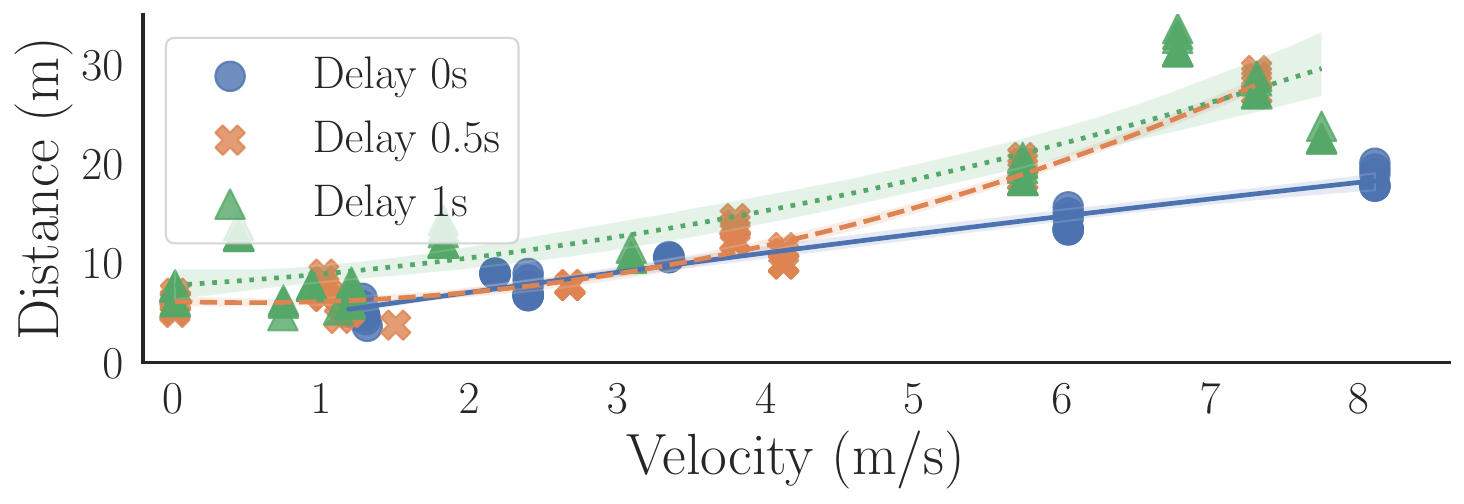}
    \caption{Shield activation for pedestrian scenario.}
    \label{fig:carlacross}
\end{figure}

\paragraph{Results.}
In Figure~\ref{fig:carlacross} we plot, 
for each interference of the shield, 
the distance from the pedestrian and the speed of the car at which the shield interferes. 
Since pedestrians are modelled in such a way that they are able to move towards the car,
the shield has to consider actual pedestrian positions closer
to the car than observed due to the delays in sensing the pedestrian.
The larger the delay, the more uncertainty the
shield has about the current position of the pedestrian and the earlier the shield initiates braking.
The synthesis times are given in Table~\ref{tab:computingtimes}.
In our experiments, the game enters occasionally states with empty strategy
due to discretization errors. 
However, the safety specification was never violated.

\section{Experiments - Shielding in a Gridworld}

\begin{figure}
 \centering
 \includegraphics[height=4.2cm]{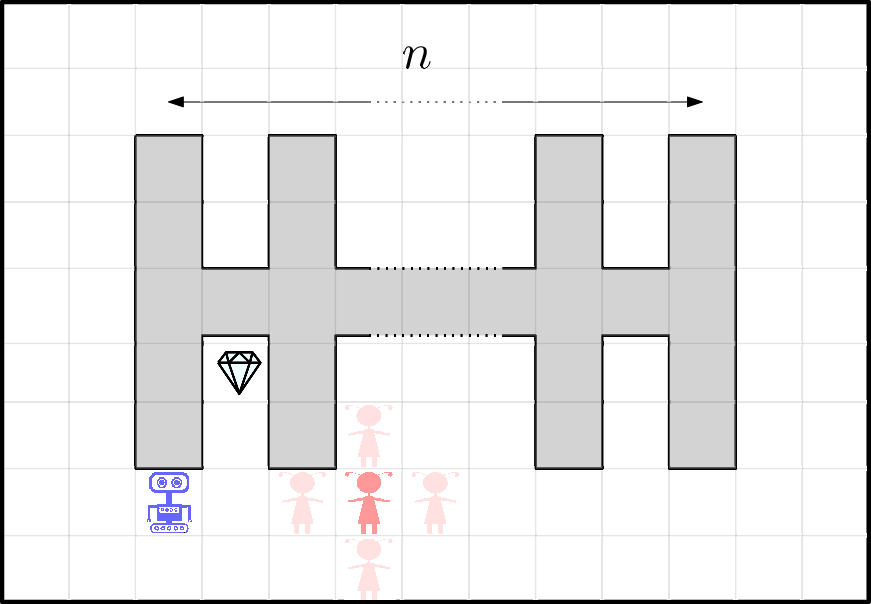}
 \caption{Gridworld with possible states after delay $\delta=1$.}
 \label{fig:exp3left}
\end{figure}

\paragraph{Setting.}
Our final case study is an extension of the one from~\cite{Chen2020IndecisionAD}.
Figure~\ref{fig:exp3left} illustrates a grid world of size $3n+4\times 9$,
where the width is parameterized
by the number of pairs of dead-ends $ n$.
There are two actors that operate in the grid world: 
a robot (controlled by the agent) and 
a kid (controlled by the environment).
The safety specification requires the robot to avoid any collision with the kid.

\paragraph{Game Graph.}
The game graph 
encoding the relevant safety dynamics for the grid world is $\mathcal G =\langle S, \mathcal E \rangle$. 
The states encode the position of both the robot and the kid. 
Thus a state is of the form $(x_0,y_0,x_1,y_1)$, 
where $(x_0,y_0)$ is the position of the robot and 
$(x_1, y_1)$ is the position of the kid.
Input letters modify the position of the kid $(x_0,y_0)$, 
while action letters modify the position of the robot $(x_1,y_1)$.
At every time step, the kid can move 1 step in each direction.
The robot can move zero, one or two steps in each direction, and can also perform three-step L-shaped moves.
Any illegal transition (those that would go out of boundaries or clash with the grey region depicted in Figure~\ref{fig:exp3left}) is changed to
$\mathtt{N}$ (``no move'').

\begin{table}[b]
\begin{tabular}{llllll}
\multicolumn{2}{c}{Delay (steps)}                                & 0    & 1   & 2     & 3     \\ \hline 
\multirow{2}{*}{Score}                & Robust.    & 42.5    & 34.3  & 31.5   & 26.8   \\ \cline{2-6} 
                                      & Control.   & 41.3  & 33.9 & 31.8      & 27.5    \\ \hline
\multirow{2}{*}{Interventions}        & Robust.    & 90.9   & 107.5  & 114.1 & 122.0  \\ \cline{2-6} 
                                      & Control. & 85.0 & 95.9  & 106.9    & 122.7
\end{tabular}
\caption{Performance of different shielding strategies.}
\label{tab:exp3}
\end{table}

\paragraph{Results: Interference Rates.} To evaluate the interference of the shields during runtime, we implemented a robot with the goal to collect treasures that are placed at random positions in a grid world with $4$ dead ends. At any time step, there is one treasure placed in the grid world. As soon as this treasure is collected, the next treasure spawns at a random position. Collecting a treasure rewards the agent with $+1$.)
The kid is implemented such that it chases the robot in a stochastic way.

Table~\ref{tab:exp3} shows for delays $\delta\in\{0,1,2,3\}$
(1) the score obtained by the robot and
(2) the number of times the shield had to intervene
on plays of $2000$ time steps. 
Since both the robot and the kid are implemented with stochastic behaviour,
each data point in the table is the average of 100 plays.

The results show that the agent's score decreases with the delay, as expected.
Since the shield has more uncertainty about the current position of the kid,
it enforces a larger distance between the current position of the robot and the last observed position of the kid. 
For the same reason, the shields need to interfere more frequently 
with increasing delays.
Additionally, we compared the corrective actions 
chosen by shields that maximize controllability with the actions chosen by shields that maximize robustness. We noticed that in most states, both shields pick the same corrective action,
leading to similar results. 

\begin{figure}[t]
         \centering
         \includegraphics[width=0.458\textwidth]{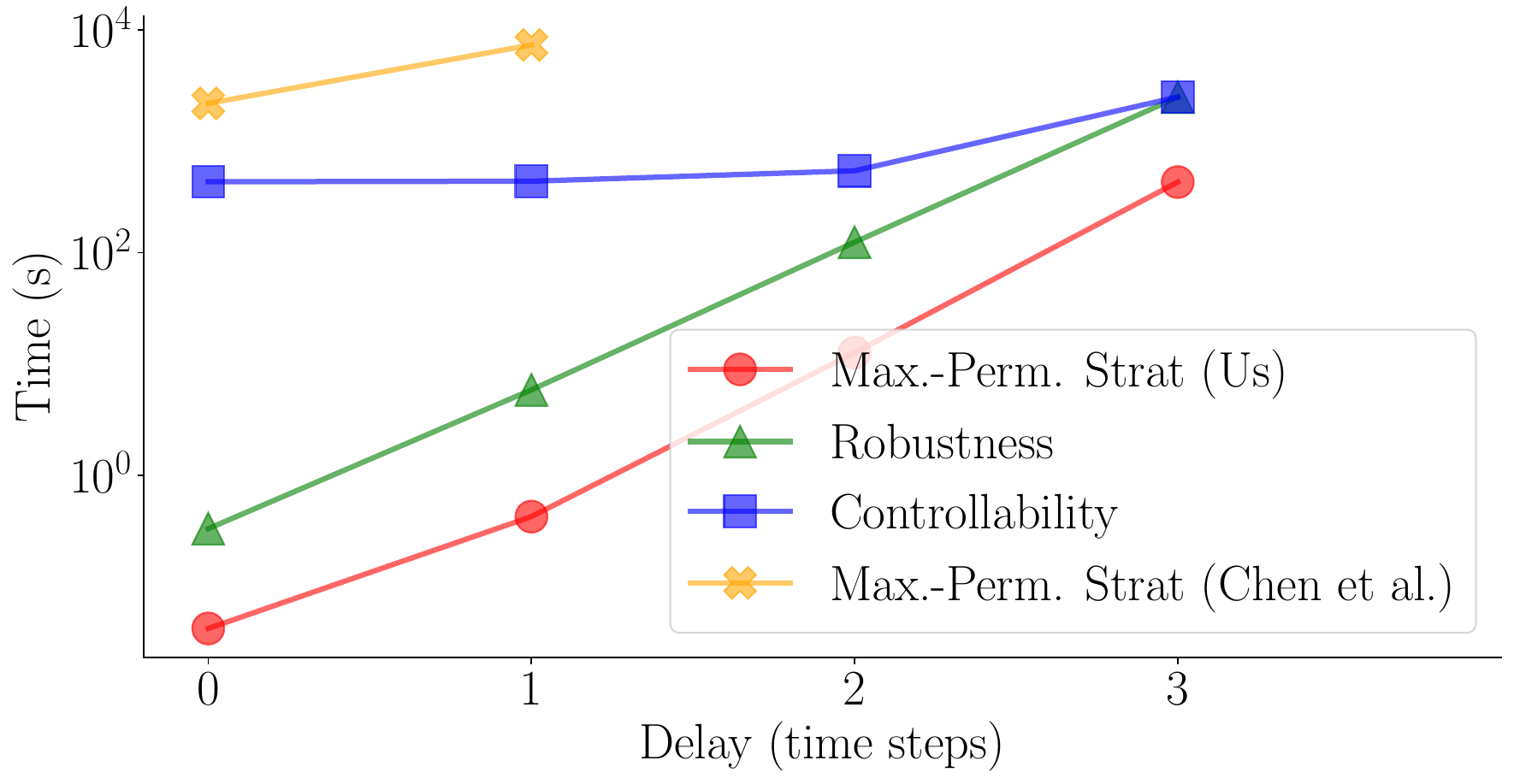}
         \caption{Synthesis times for a fixed number of 4 dead ends.}
         \label{fig:computation_timeleft}
 \end{figure}

\paragraph{Results: Synthesis Times.}
Figure~\ref{fig:computation_timeleft} depicts synthesis times against delays
for shields maximizing robustness $\phi_{r}$ (\inlinegraphics{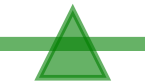}),
and controllability $\phi_{c}$ (\inlinegraphics{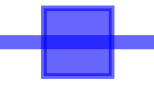}), respectively.
To compare with a baseline,
we also include the cost of computing 
the maximally-permissive strategy in the delayed safety game 
for our implementation (\inlinegraphics{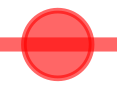})
and the implementation of~\cite{Chen2020IndecisionAD}
(\inlinegraphics{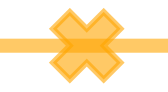}).
The improvement of our method compared to the baseline 
results from a faster implementation in C++,
with only minor algorithmic reasons.
The cut-off value for controllability is set to $\delta_{\max} = 3$.
Since the cost for computing shields grows exponentially with $\delta$,
the synthesis times for shields maximizing robustness grow exponentially. 
This effect does not show for shields maximizing controllability, 
as they always compute the maximally-permissive strategy until delay $\delta_{\max}$ irrespective of the particular delay $\delta$.

\section{Conclusion} 
\label{sec:conclusion}

We propose a new synthesis approach to construct shields 
that are able to guarantee safety under delays in the input data.
We introduce two shielding strategies that are specifically targeted to minimize 
shield interference.
We demonstrate the applicability of our approach in complex applications such as autonomous driving. 
In future work, we want to develop shields that are both resilient to delays and able to achieve high performance in probabilistic environments. 
Computing delay-resilient games for continuous time by solving timed safety games 
is also a promising direction for further research.

\section{Acknowledgments}
This work has received funding from the European Union’s Horizon 2020 research
and innovation programme under grant agreement N$^\circ$ 956123 - FOCETA.
It also received funding from  Deutsche Forschungsgemeinschaft under grant no.\ 
DFG FR 2715/5-1 ``Konfliktresolution und kausale Inferenz mittels integrierter sozio-technischer Modellbildung'', and by the State of Lower Saxony within the Zukunftslabor Mobilit\"at.
This work was also supported in part by the State Government of Styria, Austria – Department Zukunftsfonds Steiermark.

\bibliography{aaai23}

\end{document}